\documentclass[10pt,twocolumn,letterpaper]{article}

\usepackage{cvpr}
\usepackage{times}
\usepackage{epsfig}
\usepackage{graphicx}
\usepackage{amsmath}
\usepackage{amssymb}

\usepackage{epstopdf}
\usepackage{url}

\usepackage{times}

\usepackage{url}

\usepackage{amsfonts}

\usepackage{graphicx}

\usepackage{parskip}

\usepackage{multirow}

\usepackage{psfrag}

\usepackage{verbatim}

\usepackage{wrapfig}

\usepackage{paralist}

\usepackage{enumitem}
\setdescription{leftmargin=0.25cm}

\usepackage[usenames,dvipsnames]{color}

\newcounter{mycounter}

\usepackage[]{algorithm2e}
\makeatletter
\newcommand{\Removelatexerror}{\let\@latex@error\@gobble}
\makeatother
\newcommand{\myalgorithm}{%
\SetInd{0.5em}{0.5em}
\let\oldnl\nl
\newcommand{\nonl}{\renewcommand{\nl}{\let\nl\oldnl}}
\newcommand{\pushline}{\Indp}
\newcommand{\popline}{\Indm\dosemic}
\begingroup
\Removelatexerror 
\begin{algorithm*}[H]
\end{algorithm*}
\endgroup}

\usepackage{fixltx2e}

\newcommand{\bR}{\mathbf{R}}

\newcommand{\bT}{\mathbf{T}}

\newcommand{\btheta}{\mbox{\boldmath$\theta$}}

\def\I{{\cal I}}

\def\rev#1{{#1}}

\makeatletter
\renewcommand{\paragraph}{%
  \@startsection{paragraph}{4}%
  {\z@}{0.25ex \@plus 1ex \@minus .2ex}{-1em}%
  {\normalfont\normalsize\bfseries}%
}
\makeatother

\usepackage[pagebackref=true,breaklinks=true,letterpaper=true,colorlinks,bookmarks=false]{hyperref}

\cvprfinalcopy 

\def\cvprPaperID{1431} 

\ifcvprfinal\fi
\begin{document}

\title{3D Shape Segmentation with Projective Convolutional Networks}

\author{ 
        Evangelos Kalogerakis$^1$
    \and
        Melinos Averkiou$^2$
    \and
        Subhransu Maji$^1$
    \and
        Siddhartha Chaudhuri$^3$
     \and
     \\
    $^1$University of Massachusetts Amherst \,\,\,\,\,\, $^2$University of Cyprus \,\,\,\,\,\, $^3$IIT Bombay
}

\maketitle

\begin{abstract} \vspace{-0.2in}
This paper introduces a deep architecture for segmenting 3D objects into their labeled semantic parts. Our architecture  combines image-based Fully Convolutional Networks (FCNs) and surface-based Conditional Random Fields (CRFs) to yield coherent segmentations of 3D shapes. The image-based FCNs are used for efficient view-based reasoning about 3D object parts. Through a special projection layer, FCN outputs are effectively aggregated across multiple views and scales, then are projected onto the 3D object surfaces. Finally, a surface-based CRF combines the projected outputs with geometric consistency cues to yield coherent segmentations. The whole architecture (multi-view FCNs and CRF) is trained end-to-end. Our approach significantly outperforms the existing state-of-the-art methods in the currently largest segmentation benchmark (ShapeNet). Finally, we demonstrate promising segmentation results on noisy 3D shapes acquired from consumer-grade depth cameras.
\end{abstract}
\vspace{-0.2in}

\vspace{-2mm}
\section{Introduction}
\vspace{-2mm}
In recent years there has been an explosion of 3D shape data on the web. In addition to the increasing number of community-curated CAD models, depth sensors deployed on a wide range of platforms are able to acquire  3D geometric representations
of objects  in the form of polygon meshes or point clouds. Although there have been significant advances in analyzing color images, in particular through deep networks, existing semantic reasoning techniques for 3D\ geometric shape data mostly rely on heuristic processing stages and hand-tuned geometric descriptors.


Our work focuses on the task of segmenting 3D shapes into labeled semantic parts. Compositional part-based reasoning for 3D shapes has been shown to be effective for a large number of vision, robotics\ and virtual reality applications, such as cross-modal analysis of 3D shapes and color images~\cite{zhou2016learning,Huang2015rec},  skeletal tracking \cite{shotton2011real}, objection detection in images \cite{Fidler2012objectdetection,lim2014objectdetection,pepik15objectdetection}, 3D object reconstruction from images and line drawings~\cite{Xue2012rec,Huang2015rec,huang2016shape}, interactive assembly-based 3D modeling~\cite{Chaudhuri2013assembly,Chaudhuri2013attribit}, generating
3D shapes from a small number of examples \cite{kalogerakis2012probabilistic}, style transfer between 3D objects \cite{Lun2016styletransfer}, robot navigation and grasping \cite{Rusu2013robotics,Chiu2010grasping},
to name a few.

The shape segmentation task, while fundamental, is challenging because of the variety and ambiguity of shape parts that must be assigned the same semantic label; because accurately detecting boundaries between parts can involve extremely subtle cues; because local and global features must be jointly examined; and because the analysis must be robust to noise and undersampling.

We propose a deep architecture for segmenting and labeling 3D shapes that simply and effectively addresses these challenges, and significantly outperforms\ prior methods. The key insights of our technique are to repurpose image-based
deep networks
for view-based reasoning, and aggregate
their outputs onto the surface representation of the shape in a geometrically consistent manner.
\rev{ We make no geometric, topological or orientation assumptions about the shape, nor    exploit any hand-tuned geometric descriptors.}

\begin{figure*}[t!]
  \begin{center}
  \includegraphics[width=1.02\linewidth]{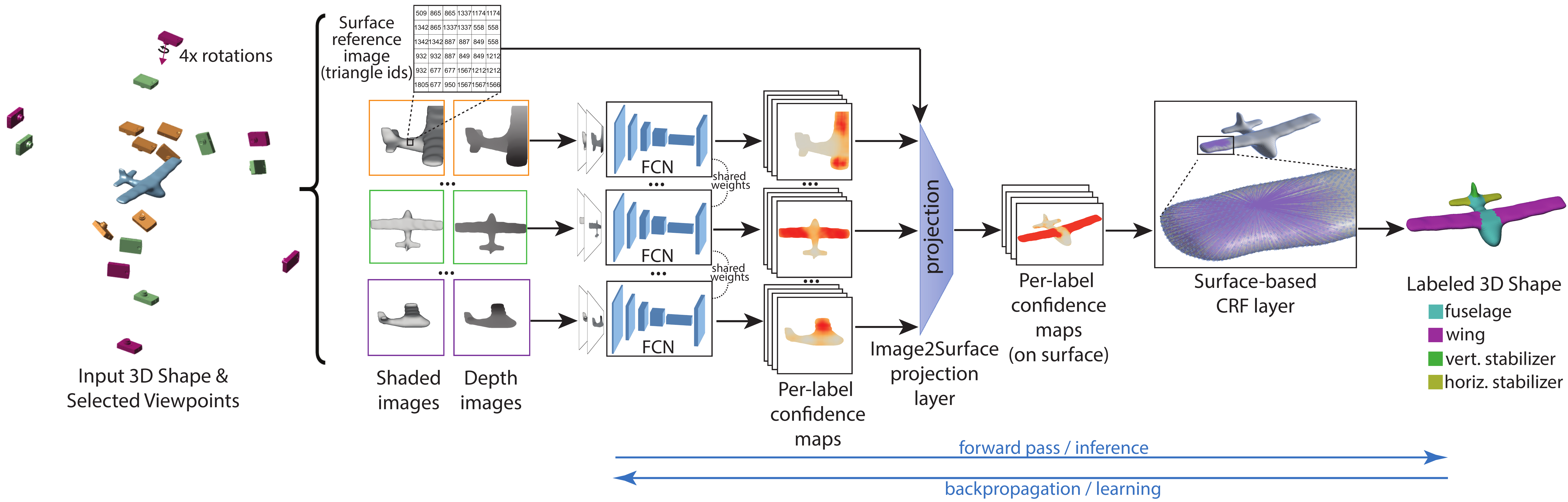}
  \caption{\label{fig:mvfcn-archictecture} Pipeline and architecture of our method for 3D shape segmentation and labeling. Given an input shape,  a set of viewpoints are computed at different scales such that the viewed shape surface is maximally covered (left). Shaded and depth images from these viewpoints are processed through our architecture (here we show images for three viewpoints, corresponding to 3 different scales). Our architecture employs image-based Fully Convolutional Network (FCN) modules with shared parameters to process the input images. The modules output image-based part label confidences per view. Here we show confidence maps for the wing label (the redder the color, the higher the confidence). The confidences are aggregated and projected on the shape surface through a special projection layer. Then they are further processed through a surface-based CRF that promotes consistent labeling of the entire surface (right). }
\vskip -4mm
  \label{fig:architecture}
  \end{center}
\end{figure*}

Our view-based approach is motivated by the success of deep networks on image segmentation tasks.  Using rendered shapes lets us initialize our network with layers that have been trained on large image datasets,  allowing better generalization. Since images depict shapes of photographed objects (along with texture), we expect such pre-trained layers to already encode some information about parts and their relationships. Recent work on view-based 3D shape classification~\cite{su2015multi,qi2016volumetric} and RGB-D recognition~\cite{gupta2014learning,song2015deep} have shown the benefits of transferring learned representations from color images to geometric and depth data.

A view-based approach to 3D shape segmentation must overcome several technical obstacles. \rev{First, views must be selected such that they together cover the shape surface as much as possible and minimize occlusions.} Second, shape parts can be visible in more than one view, thus our method must effectively consolidate information across multiple views. Third, we must guarantee that the segmentation is complete and coherent. This means all the surface area, including  any heavily occluded portions, should be labeled, and neighboring surface areas should likely have the same label unless separated by a strong boundary feature.

Our approach, shown in Figure~\ref{fig:mvfcn-archictecture}, systematically addresses these difficulties using a single feed-forward network. Given a raw 3D polygon mesh as input, our method generates a set of images from multiple views that are automatically selected for optimal surface coverage. These images are fed into the network, which outputs  confidence maps per part via image processing layers. The confidence maps are fused and projected onto the shape surface representation through a  projection layer. Finally, our architecture incorporates a surface-based Conditional Random Field (CRF) layer that promotes consistent labeling of the entire surface. The whole network, including the CRF, is trained  in an end-to-end manner to achieve optimal performance.

\rev{Our main contribution is the introduction of a deep architecture
 for compositional part-based reasoning on 3D\ shape representations
without the use of hand-engineered geometry processing stages or hand-tuned descriptors.
We demonstrate significant improvements over the state-of-the-art. For complex objects, such as  aircraft, motor vehicles, and furniture, our method increases part labeling accuracy by a remarkable $\mbox{$\sim$8\%}$  over the state of the art on the currently largest 3D\ shape segmentation dataset.}

\vspace{-2mm}
\section{Related work}
\vspace{-3mm}
Our work is related to learning methods for segmentation of images (including RGB-D data) and 3D\ shapes.
\vspace{-2mm}
\paragraph{Image-based segmentation.} There is a vast literature on segmenting images into objects and their parts. Most recent techniques are based on variants of random forest classifiers or convolutional networks. An example of the former is the remarkably fast and accurate human-pose estimator that uses depth data from Kinect sensors for labeling human parts~\cite{shotton2011real}. Our work builds on the success of convolutional networks for material segmentation, scene labeling, and object part-labeling tasks. These approaches use image classification networks repurposed for dense image labeling, commonly a fully-convolutional network (FCN)~\cite{long2015fully}, to obtain an initial labeling. Several strategies for improving these initial estimates have been proposed including techniques based on top-down region-based reasoning~\cite{cimpoi2015deep,hariharan2014simultaneous}, CRFs~\cite{chen14semantic,lin2015efficient}, atrous convolutional layers~\cite{chen14semantic,yu2016dilation}, deconvolutional layers~\cite{noh2015learning}, recurrent networks~\cite{zheng2015conditional}, or a multi-scale analysis~\cite{mostajabi2015feedforward,hariharan2015hypercolumns}.
Several works~\cite{lai2011large,blum2012learned,bo2013unsupervised} have also focused on learning feature representations from RGB-D data (e.g. those captured using a Kinect sensor) for object-level recognition and detection in scenes. Recently, Gupta et al.~\cite{gupta2014learning} showed that image-based networks can be repurposed for extracting depth representations for object detection and  segmentation. Recent works~\cite{gupta2013perceptual,song2015sun,hazirbasma2016fusenet} have  applied a similar strategy for  indoor scene recognition tasks.

In contrast to the above methods, our work aims to segment geometric representations of 3D objects, in the form of polygon meshes, created through 3D modeling tools or reconstruction techniques. The 3D models of these objects often do not contain texture or color information. Segmenting these 3D objects into parts requires architectures that are capable of operating on their geometric representations.
\vspace{-5.6mm}
\paragraph{Learning 3D shape representations from images.} \rev{A few recent methods attempt to learn volumetric  representations of shapes from images via convolutional networks that employ  special layers to model shape projections onto images \cite{Yan16,Rezende16}. Alternatively, mesh-based representations can also be learned from images by assuming a fixed number of mesh vertices \cite{Rezende16}. In contrast to these works, our architecture  discriminatively learns view-based shape representations along with a surface-based CRF such that the view projections match an input surface signal (part labels). Our 3D-2D projection mechanism  is differentiable, parameter-free, and sparse, since it operates only on the shape surface rather than its volume. In contrast to the  mesh representations of \cite{Rezende16}, we do not assume that meshes have a fixed number of vertices, which does not hold true for general 3D models. Our method is more related to methods that learn view-based shape representations  ~\cite{su2015multi,qi2016volumetric}. However, these methods only learn global  representations  for shape classification and rely on fixed sets of views. Our method instead learns view-based shape representations for part-based reasoning through adaptively selected views. It also uses a CRF to resolve inconsistencies or missing surface information in the view representations.}

\vspace{-2mm}
\paragraph{3D geometric shape segmentation.} The most common learning-based approach to shape segmentation is to assign part labels to geometric elements of the shape representation, such as polygons, points, or patches \cite{kai2015star}. This is often done through various processing stages: first, hand-engineered geometric descriptors of these elements are extracted (e.g. surface curvature, shape diameter, local histograms of point or normal distributions, surface eigenfunctions, etc.); then, a clustering method or classifier infers part labels for elements based on their descriptors; and finally (optionally) a separate graph cuts step is employed to smooth out the surface labeling
~\cite{Kalogerakis2010,Shapira2010,Sidi2011,Hu2012,Yumer2014}.
Recently, a convolutional network has been proposed as an alternative   element classifier \cite{Guo2015}, yet it operates on hand-engineered geometric descriptors organized in a 2D matrix lacking spatially coherent structure for conventional convolution. Another variant is to use two-layer networks which transform the input by randomized kernels, in the form of so-called ``Extreme Learning Machines'' \cite{Xie2015}, but these offer no better performance than standard shallow classifiers.

Other approaches segment shapes by employing non-rigid alignment steps through deformable part templates \cite{Kim2013,Huang2015deeplearningsurfaces}, or transfer labels through surface correspondences and functional maps between 3D shapes  \cite{vanKaick2011,Huang2011,Wang2012,Kim2013,Huang2014}.
These correspondence and alignment methods  rely  on hand-engineered geometric descriptors and deformation steps. Wang et al.~\cite{Wang2013} segment 3D shapes by warping and matching binary images of their projected views with segmented 2D images through Hausdorff distances. However, the matching procedure is hand-tuned, while potentially useful surface information, such as depth and normals, is ignored.



In contrast to all the above approaches, we propose a view-based deep architecture for shape segmentation with four main advantages. First, our architecture adopts image processing layers learned on large-scale image datasets, which are orders of magnitude larger than existing 3D datasets. As we show in this work, the deep stack of several layers extracts feature representations that can be successfully adapted to the task of shape segmentation. We note that such transfer has also been observed recently for shape recognition~\cite{su2015multi,qi2016volumetric}. Second, our architecture produces shape segmentations without the use of hand-engineered geometric descriptors or processing stages that are prone to degeneracies in the shape representation (i.e. surface noise, sampling artifacts, irregular mesh tesselation, mesh degeneracies, and so on). Third, we employ adaptive viewpoint selection to effectively capture all surface parts for analysis. Finally, our architecture is trained end-to-end, including all image and surface processing stages. As a result of these contributions, our method achieves better performance than prior work on big and complex datasets by a large margin.

\vspace{-2mm}
\section{Method}
\vspace{-3mm}
Given an input 3D shape, the goal of our method is to segment it into labeled parts.
We designed a projective convolutional network to this end. Our network architecture is visualized in Figure \ref{fig:architecture}. It takes as input a set of images from multiple views optimized for maximal surface coverage; extracts part-based confidence maps through image processing layers (pre-trained on large image datasets); combines and projects these maps onto the surface through a  projection layer, and finally incorporates a surface-based Conditional Random Field (CRF) that favors coherent labeling of the input surface. The whole network, including the CRF, is trained end-to-end. In the following sections, we discuss the input to our network, its layers, and the training procedure.
\paragraph{Input.} The input to our algorithm is a 3D shape represented as a polygon mesh. As a preprocessing step, the shape surface is sampled with uniformly distributed points ($1024$ in our implementation). Our algorithm first determines an overcomplete collection of viewpoints such that nearly every point of the surface is visible from at least $K$ viewpoints (in our implementation, $K=3$). For each sampled surface point, we place viewpoints at different distances from it along its surface normal (distances are set to $0.5$, $1.0$ and $1.5$ of the shape's bounding sphere radius). In this manner, the surface is depicted at different scales (Figure \ref{fig:architecture}, left). We then determine a compact set of informative viewpoints that maximally cover the shape surface. For each viewpoint, the shape is rasterized under a perspective projection to a binary image, where we associate every ``on'' pixel with the sampled surface point closest to it. The coverage of the viewpoint is measured as the fraction of surface points visible from it, estimated by aggregating surface point references from the image. For each of the  scales (camera distances), the viewpoint with largest coverage is inserted into a list. We then re-estimate coverages at this scale, omitting points already covered by the selected viewpoint, and the viewpoint with the next largest coverage is added to the list. The process is repeated until all surface points are covered at this scale. In our experiments, with man-made shapes and at our selected scales, approximately $20$ viewpoints were enough to cover the vast majority of the surface area per scale.

After determining our viewpoint collection, we render the shape to shaded images and depth images. For each viewpoint, we place a camera pointing towards the surface point used to generate that viewpoint, and rotate its up-vector $4$ times at $90$ degree intervals (i.e, we use 4 in-plane rotations). For each of these $4$ camera rotations, we render a shaded, greyscale \mbox{$512 \times 512$} image using a typical computer graphics shader (Phong reflection model \cite{phong1975}) and a depth image, which are concatenated into a single two-channel image. These images are fed as input to the image processing\ module (FCN) of our network, described below. \rev{We found that both shaded and depth images are useful inputs. In early experiments,  labeling accuracy dropped $2.5\%$ using depth alone.  This might be attributed to the more ``photo-realistic'' appearance of shaded images, which better match  the statistics of real images  used to pretrain our architecture. We note that shaded images  directly encode surface normals relative to  view direction (shading is computed from the  angle between normals and view direction).}

In addition to the shaded and depth images, for each selected camera setting, we rasterize the shape into another image where each pixel stores the ID of the polygon whose projection is closest to the pixel center. These images, which we call ``surface reference'' images, are fed into the ``projection layer'' of our network (Figure \ref{fig:architecture}).
\vspace{-1mm}
\paragraph{FCN module.} The  two-channel images produced in the previous step are processed through identical image-based Fully-Connected Network (FCN) modules (Figure \ref{fig:architecture}). Each FCN module outputs $L$ confidence maps of size \mbox{$512 \times 512$} per each input image, where $L$ is the number of part labels. Specifically, in our implementation we employ the FCN architecture suggested in \cite{yu2016dilation}, which adopted the VGG-16 network \cite{simonyan2015vgg} for dense prediction by removing its two last pooling and striding layers, and using dilated convolutions. We perform two additional modifications to this FCN architecture. First, since our input is a 2-channel image, we use 2-channel \mbox{$3 \times 3$} filters instead of 3-channel (BGR) ones. We also adapted these filters to handle greyscale rather than color images during our training procedure. Second, we modified the output  of the original FCN module. The original FCN outputs $L$ confidence maps of size \mbox{$64 \times 64$}. These are then converted into $L$ probability maps through a softmax operation. Instead, we upsample the confidence maps to size \mbox{$512 \times 512$} through a transpose convolutional (``deconvolution'') layer with learned parameters and stride $8$. The confidences are later converted into probabilities through our CRF layer.
\paragraph{Image2Surface projection layer.} The goal of this layer is to aggregate the confidence maps across multiple views, and project the result back onto the 3D surface.  We note that both the locations and the number of optimal viewpoints can vary from shape to shape, and they are not ordered in any manner. Even if the optimal viewpoints were the same for different shapes, the views would still not necessarily be ordered, since we do not assume that shapes are  oriented consistently. As a result, the projection layer should be invariant to the input image ordering. Given $M_s$ input images of an input shape $s$, the $L$ confidence maps extracted from the FCN module are stacked into a \mbox{$M_s \times 512 \times 512 \times L$} image. The projection layer takes as input this 4D image. In addition, it takes as input the surface reference (polygon ID) images, also stacked into a 3D \mbox{$M_s \times 512 \times 512$} image. The layer outputs a \mbox{$F_s \times L$} array, where $F_s$ is the number of polygons of the shape $s$. The projection is done through a view-pooling operation. For each surface polygon $f$ and part category label $l$, we assign a confidence $P(f,l)$ equal to the maximum label confidence across all pixels and input images that map to that polygon according to the surface reference images. Mathematically, this projection operation is formulated as:
\vspace{-1mm}
\begin{equation}
\tilde{C}(f,l) = \max\limits_{\substack{\forall m,i,j:\\I(m,i,j)=f}} C(m,i,j,l)
\vspace{-1mm}
\end{equation}
where $C(m,i,j,l)$ is the confidence of label $l$ at pixel $(i,j)$ of image $m$; $I(m,i,j)$ stores the polygon ID at pixel $(i,j)$ of the corresponding reference image $m$; and $\tilde{C}(f,l)$ is the output confidence of label $l$ at polygon $f$. We note that the surface reference images omit polygon references at and near the shape silhouette, since an excessively large, nearly occluded, portion of the surface tends to be mapped onto the silhouette, thus the projection becomes unreliable there. Instead of using the max operator, an alternative aggregation strategy would be to use the average instead of the maximum, but we observed that this results in a slightly lower performance (about $1\%$ in our experiments).

\begin{figure*}
  \begin{center}
  \includegraphics[width=\linewidth]{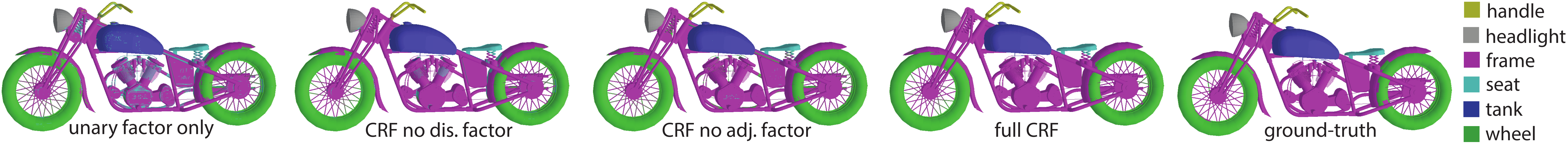}
\vskip -1mm
  \caption{\label{fig:crfanalysis} Labeled segmentation results for alternative versions of our CRF (best viewed in color).}
  \end{center}
\vskip -5mm
\end{figure*}

\textbf{Surface CRF.} Some small surface areas may be highly occluded and hence unobserved by any of the selected viewpoints, or not included in any of the reference images. For any such polygons, the label confidences are  set to zero. The rest of the surface should propagate label confidences to these polygons. In addition, due to upsampling in the FCN module, there might be bleeding across surface convexities or concavities that are likely to be segmentation boundaries.

We define a CRF operating on the surface representation to deal with the above issues. Specifically, each polygon $f$ is assigned a random variable $R_f$ representing its label. The CRF includes a unary factor for each such variable, which is set according to the confidences produced in the projection layer: $\phi_{\textrm{unary}}(R_f=l) = \exp(\tilde{C}(f,l))$. The CRF also encodes pairwise interactions between these variables based on surface proximity and curvature. For each pair of neighboring polygons $(f,f')$, we define a factor that favors the same label for polygons which share normals (e.g. on a flat surface), and different labels otherwise. Given the  angle  $\omega_{f,f'}$ between their normals ($\omega_{f,f'}$ is divided by $\pi$ to map it between $[0,1]$), the factor is defined as follows:
\begin{eqnarray}
\small
\phi_{adj}(R_f\!=\!l,\!R_{f'}\!=\!l')\!=\!
\left\{
\begin{array}{l}
\!\!\!\!
\exp \!\Big(\!\!\!-\!w_{\textrm{adj}} \!\cdot\! w_{l,l'} \!\cdot\! \omega_{f,f'}^2 \! \Big), \;\;\;\;\;\;\;\;\;l\!=\!l' \\
\!\!\!\!
\exp \!\Big(\!\!\! -\! w_{\textrm{adj}} \!\cdot\! w_{l,l'}  \!\cdot\!  (1 \!-\! \omega_{f,f'} ^2)\!\Big),\;l \!\ne\! l'\\
\end{array} \right.
\nonumber
\end{eqnarray}
where $w_{\textrm{adj}}$ and $w_{l,l'}$ are learned factor- and label-dependent weights.
We also define factors that favor similar labels for polygons $f$, $f'$ which are spatially close to each other according to the geodesic distance $d_{f,f'}$ between them. These factors are defined for pairs of polygons whose geodesic distance is less than 10\% of the bounding sphere radius in our implementation. This makes our CRF relatively dense and more sensitive to long-range interactions between surface variables. We note that for small meshes or point clouds, all pairs could be considered instead. The geodesic distance-based factors are defined as follows:
\begin{eqnarray}
\small
\phi_{dist}(R_f\!=\!l,\!R_{f'}\!=\!l')\!=\!\!
\left\{
\begin{array}{l}
\!\!\!\!
\exp \!\Big(\!\!\! -\! w_{\textrm{dist}} \!\cdot\! w_{l,l'} \!\cdot\! d_{f,f'}^2 \! \Big),
\;\;\;\;\;\;\;\;\;\;l\!=\!l' \\
\!\!\!\!
\exp \!\Big(\!\!\! -\! w_{\textrm{dist}} \!\cdot\! w_{l,l'}  \!\cdot\!  (1-d_{f,f'}^2 )\!\Big),\;l \!\ne\! l'
\end{array} \right.
\nonumber
\end{eqnarray}
where the factor-dependent weight $w_{\textrm{dist}}$ and label-dependent weights $w_{l,l'}$ are learned parameters, and $d_{f,f'}$ represents the geodesic distance between $f$ and $f'$ . Distances are normalized to $[0,1]$.

Based on  the above factors, our CRF is defined over all surface random variables $\bR_s=\{R_1,R_2, \dots, R_{F_s}\}$ of the shape $s$ as follows:
\begin{equation}
\small
P(\bR_s)\!=\!\frac{1}{Z_s}\! \prod\limits_f \phi_{\textrm{unary}}(R_f)\!\!
\prod\limits_{\textrm{adj}\,f,f'} \! \phi_{\textrm{adj}}(R_f,R_{f'}) \!\!
\prod\limits_{\textrm{}\,f,f'} \! \phi_{\textrm{dist}}(R_f,R_{f'})
\end{equation}
where $Z_s$ is a normalization constant. Exact inference is intractable, thus we resort to mean-field inference to approximate the most likely joint assignment to all  random variables as well as their marginal probabilities. \rev{Our mean-field approximation  uses distributions over single variables as messages (i.e. the posterior is approximated in a fully factorized form -- see Algorithm 11.7 of \cite{kf-pgmpt-09})}. Figure \ref{fig:crfanalysis} shows how segmentation results degrade for alternative versions of our CRF, and when the unary term  is used alone.

\vspace{-1.5mm}
\paragraph{Training procedure.} The FCN module is initialized with filters pre-trained on  image processing tasks \cite{yu2016dilation}.  Since the input to our network are rendered grayscale (colorless) images, we average the BGR channel weights of the pre-trained filters of the first convolutional layer, i.e. the \mbox{$3 \times 3 \times 3$} filters are converted to color-insensitive \mbox{$3 \times 3 \times 1$} filters. Then, we replicate the weights twice to yield \mbox{$3 \times 3 \times 2$} filters that can accept our 2-channel input images. The CRF weights are initialized to 1.

Given an input training dataset $S$ of 3D shapes, we first generate their depth, shaded, and reference images using our rendering procedure. Then, our algorithm fine-tunes the FCN module filter parameters $\btheta$ and learns the CRF weights $w_{\textrm{adj}},w_{\textrm{dist}}, \{w_{l,l'}\}$ to maximize their log-likelihood plus a small regularization term:
\vspace{-2mm}
\begin{equation}
L=\frac{1}{|S|}\sum\limits_{s \in S} \log P(\bR_s=\bT_s) + \lambda ||\btheta||^2
\vspace{-2mm}
\end{equation}
where $\bT_s$ are ground-truth labels per surface variable for the training shape $s$, and $\lambda$ is a regularization parameter (weight decay) set to $10^{-3}$ in our experiments. To maximize the above objective, we must  compute its gradient w.r.t. the FCN module outputs, as required for backpropagation:
\begin{equation}
\small
\frac{\partial L}{ \partial C(m,i,j,l)}\!=\!\!
\left\{ \begin{array}{l}
1-P(R_f=l) \mbox{\:\:if\:\:} l=T_f \mbox{\:and\:} I(m,i,j)=f \\
P(R_f=l)  \mbox{\:\:\:\:\:\:\:\:\:\:if\:\:} l\ne T_f \mbox{\:and\:} I(m,i,j)=f \\
0         \mbox{\:\:\:\:\:\:\:\:\:\:\:\:\:\:\:\:\:\:\:\:\:\:\:\:\:\:\:\:otherwise} \end{array} \right.
\end{equation}
Computing the gradient requires estimation of the marginal probabilities $P(R_f)$. We use mean-field inference to estimate the marginals (same inference procedure is used for training and testing). We observed that after $20$ iterations, mean-field often converges (i.e. marginals change very little). We also need to compute the gradient of the objective function w.r.t. the CRF weights. Since our CRF has the form of a log-linear model,  gradients can be easily derived.

\rev{Given the estimated gradients, we can train our network through backpropagation. Backpropagation can send error messages towards any FCN branch i.e., any input image (Figure \ref{fig:architecture}). One strategy to train our network would be to set up as many FCN branches as the largest number of rendered images across all training models. However, the number of selected viewpoints varies per model, thus  the number of rendered images per model also varies, ranging from a few tens to a few hundreds in our datasets. Maintaining hundreds of FCN\ branches would exceed the memory capacity of current GPUs.
Instead, during training, our strategy is to pick a random  subset of $24$ images per model, i.e. we keep $24$ FCN branches with shared parameters in the GPU\ memory.   For each batch, a different random subset per model is selected (i.e. no fixed  set of views used for training).  We  note  that the order of  rendered images does not matter --  our view pooling  is invariant to the input image ordering. Our training strategy is reminiscent of the DropConnect technique \cite{wan2013dropconnect}, which  tends to reduce overfitting.}

\rev{At test time all rendered images per model are used to make  predictions. The forward pass does not require all the input images to be processed at once (i.e., not all  FCN branches need to be set up). At test time, the image label confidences are sequentially projected onto the surface, which produces the same results as projecting all of them at once.}
\vspace{-2mm}
\paragraph{Implementation.}  Our network is implemented using C++ and Caffe \footnote{Our source code, results and datasets are available on the project page: \url{http://people.cs.umass.edu/kalo/papers/shapepfcn/}}. Optimization is done through stochastic gradient descent with learning rate $10^{-3}$  and momentum $0.9$. We implemented a new Image2Surface layer in Caffe for projecting image-based confidences onto the shape surface. We also created a CRF layer that handles mean-field inference during the forward pass, and estimates the required gradients during backpropagation.

\vspace{-2mm}
\section{Evaluation}
\vspace{-3mm}
\begin{table}[t]
\small
 \centering
    \begin{tabular}{@{}c@{}||@{}c@{}|@{}c@{}|@{}c@{}|@{}c@{}|@{}c@{}}
      & \,\#train/test\, & \#part& \multirow{2}{*}{\,ShapeBoost\,} & \multirow{2}{*}{\,Guo et al.\,}    & \multirow{2}{*}{\,ShapePFCN\,}\\
      &     shapes       & \,labels\,&                          &                            &   \\
    \hline
    \hline
    Airplane    & 250 / 250   & 4     & 85.8  & 87.4  & \textbf{90.3} \\
    \hline
    Bag         & 38 / 38     & 2     & 93.1  & 91.0  & \textbf{94.6} \\
    \hline
    Cap         & 27 / 28     & 2     & 85.9  & 85.7  & \textbf{94.5} \\
    \hline
    Car         & 250 / 250   & 4     & 79.5  & 80.1  & \textbf{86.7} \\
    \hline
    Chair       & 250 / 250   & 4     & 70.1  & 66.8  & \textbf{82.9} \\
    \hline
    Earphone    & 34 / 35     & 3     & 81.4  & 79.8  & \textbf{84.9} \\
    \hline
    Guitar      & 250 / 250   & 3     & 89.0  & 89.9  & \textbf{91.8 }\\
    \hline
    Knife       & 196 / 196   & 2     & 81.2  & 77.1  & \textbf{82.8} \\
    \hline
    Lamp        & 250 / 250   & 4     & 71.7  & 71.6  & \textbf{78.0} \\
    \hline
    Laptop      & 222 / 223   & 2     & 86.1  & 82.7  & \textbf{95.3 }\\
    \hline
    Motorbike   & 101 / 101   & 6     & 77.2  & 80.1  & \textbf{87.0 }\\
    \hline
    Mug         & 92 / 92     & 2     & 94.9  & 95.1  & \textbf{96.0}\\
    \hline
    Pistol      & 137 / 138   & 3     & 88.2  & 84.1  & \textbf{91.5} \\
    \hline
    Rocket      & 33 / 33     & 3     & 79.2  & 76.9  & \textbf{81.6} \\
    \hline
    Skateboard  & 76 / 76     & 3     & 91.0  & 89.6  & \textbf{91.9} \\
    \hline
    Table      & 250 / 250    & 3     & 74.5  & 77.8  & \textbf{84.8} \\
    \hline
    \end{tabular}%
  \caption{Dataset statistics and labeling accuracy per category for test shapes in ShapeNetCore (see also Table  \ref{tab:upresults_per_category} for accuracy in the case of consistent upright orientation for shapes).}
\vspace{-2mm}
  \label{tab:results_per_category}
\end{table}%

\begin{table}[t]
\small
  \centering
    \begin{tabular}{@{}c@{}|@{}c@{}|@{}c@{}|@{}c@{}}
    & \,ShapeBoost\, & \,Guo et al.\,    & \,ShapePFCN\, \\
    \hline
    \hline
    Category Avg. & 83.0 & 82.2 & \textbf{88.4} \\
    \hline
    Category Avg. (\textgreater3 labels)\, & 76.9 & 77.2 & \textbf{85.0} \\
    \hline
    Dataset  Avg. & 81.2 & 80.6 & \textbf{87.5} \\
    \hline
    Dataset  Avg. (\textgreater3 labels) & 76.8 & 76.8 & \textbf{84.7 }\\
    \hline
    \end{tabular}%
  \caption{Aggregate labeling accuracy on ShapeNetCore.}
\vspace{-5mm}
  \label{tab:results_aggregated}
\end{table}%

We now present experimental validations and analysis of our approach.

\begin{figure*}
        \begin{center}
      \includegraphics[width=\linewidth]{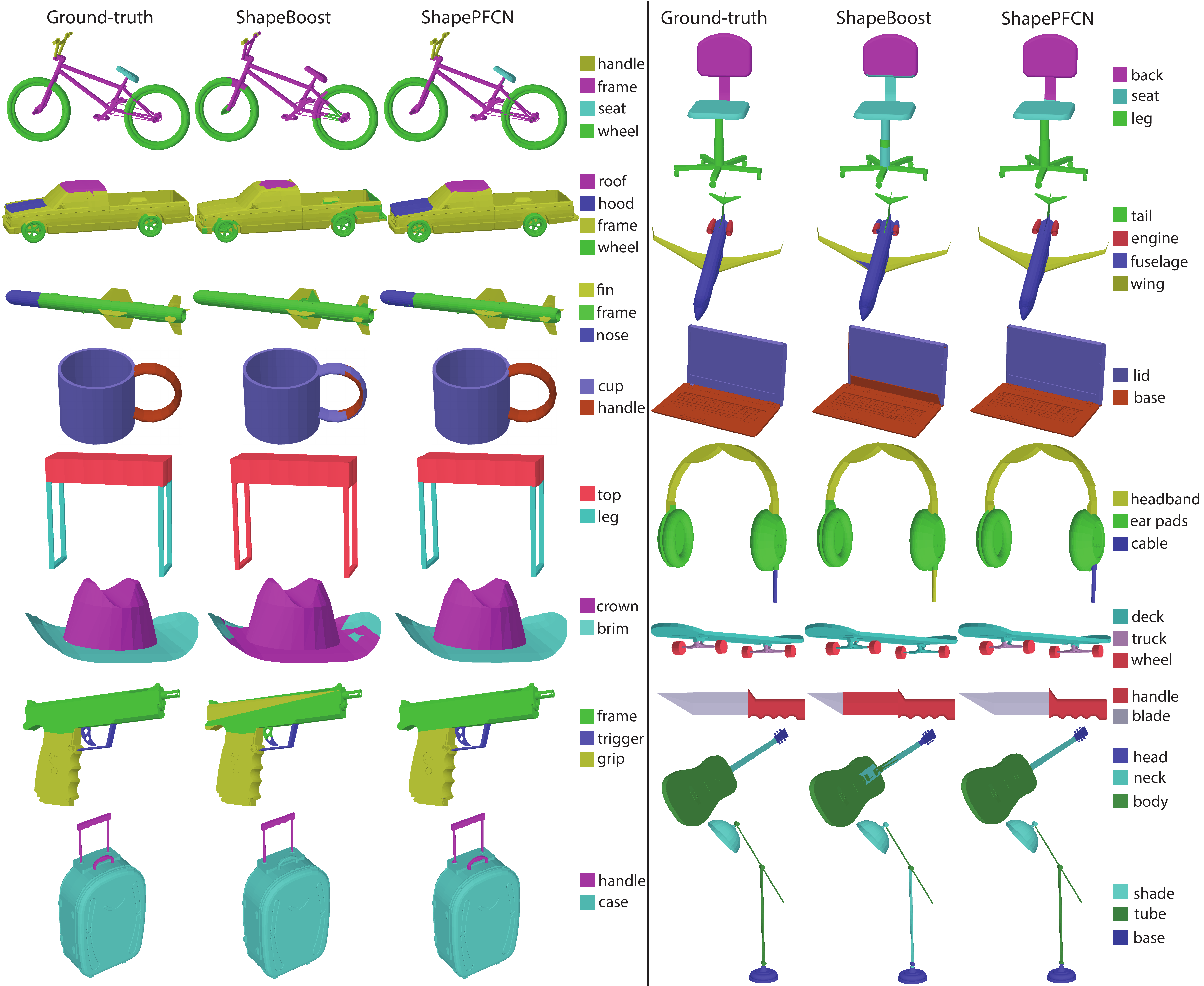}
        \vskip -1mm
        \caption{\label{fig:gallery} Ground-truth (human) labeled segmentations of ShapeNet shapes, along with segmentations produced by ShapeBoost \cite{Kalogerakis2010} and our method (ShapePFCN) for test shapes originating from the ShapeNetCore dataset (best viewed in color).         }
        \end{center}
        \vskip -4mm
\end{figure*}

\vspace{-1mm}
\textbf{Datasets.} We evaluated our method on manually-labeled segmentations available from the ShapeNetCore \cite{Yi2016}, Labeled-PSB (L-PSB) \cite{chen2009psb,Kalogerakis2010}, and COSEG datasets \cite{Wang2012}.  The dataset from ShapeNetCore currently contains 17,773 ``expert-verified'' segmentations of 3D models across $16$ categories. The 3D models of this dataset are gathered ``in the wild''. They originate from the Trimble 3D Warehouse and Yobi3D repositories, and in general are typical representatives of objects created using 3D modeling tools for diverse applications. In contrast, the PSB and COSEG datasets are much smaller. PSB contains $380$ segmented 3D models in $19$ categories ($20$ models per category), while COSEG contains $190$ segmented models in $8$ categories, plus $900$ synthetic variations of those in $3$ categories. All models in the PSB have been carefully re-meshed and reconstructed such that their mesh representation is watertight with clean topology \cite{Wang2012}, facilitating use in geometry processing applications. Most shapes in COSEG are similarly preprocessed. As the authors of the PSB benchmark note \cite{Wang2012}, many 3D models cannot be re-meshed or re-constructed due to mesh degeneracies, hence they were not included in their datasets. From this aspect, our analysis is primarily focused on the dataset from ShapeNetCore, since it is by far the largest of the three datasets; contains diverse, general-purpose 3D models; and was gathered ``in the wild''. Nevertheless, for completeness, we include comparisons with prior methods on all  datasets (ShapeNetCore, L-PSB, COSEG).
\vspace{-1mm}
\paragraph{Prior methods.} We include comparisons with: (i) ``ShapeBoost', the method described in \cite{Kalogerakis2010}, which employs graph cuts with a cascade of JointBoost classifiers for the unary term and GentleBoost for the pairwise term along with other geometric cues, and has state-of-the-art performance on the L-PSB; (ii) the recent method by Guo et al. \cite{Guo2015} which reports comparable performance on the L-PSB dataset with ShapeBoost. This method employs graph cuts with a CNN on per-face geometric descriptors (also used in ShapeBoost), plus geometric cues for the pairwise term.

Computing geometric descriptors on ShapeNetCore shapes is challenging since they are often non-manifold ``polygon soups'' (meshes with arbitrarily disconnected sets of polygons) with inconsistently oriented surface normals. Working with the original publicly available ShapeBoost implementation, we tried to make the computation of geometric descriptors and graph cuts as robust as possible. We preprocessed the meshes to correctly orient polygons (front-facing w.r.t. external viewpoints), repair connectivity (connect geometrically adjacent but topologically disconnected polygons, weld coincident vertices), and refine any excessively coarse mesh by planar subdivision of faces until it has $>$3,000 polygons. We also computed the geometric descriptors on point-sampled representations of the shape so that they are relatively invariant to tessellation artifacts. We note that neither the prior methods nor our method make any assumptions about shape orientation. No method explicitly used any pre-existing mesh subpart information manually entered by 3D modelers. Finally, we note that in the absence of a publicly available implementation, we used our own implementation of Guo et al's architecture.

\textbf{Dataset splits.} Since a standard training/test split is not publicly available for the segmented ShapeNetCore dataset, we introduced one (full list in the supplementary material). We randomly split each category into two halves, $50\%$ for training and the rest for testing. The number of 3D shapes varies significantly per category in ShapeNetCore, ranging from $66$ for Rockets to $5266$ for Tables. The computation of geometric descriptors used in prior methods is expensive, taking up to an hour for a large mesh (e.g, 50K polygons). To keep things tractable, we used $250$ randomly selected shapes for training, and $250$ randomly selected shapes for testing for categories with more than $500$ shapes. Our dataset statistics are listed in Table \ref{tab:results_per_category}.  For the much smaller PSB and COSEG datasets, we used $12$ shapes per category for training, and the rest for testing. Each of the methods below, including ours, is trained and tested separately on each shape category, following the standard practice in  prior 3D mesh segmentation literature.
 All methods used the same splits per category. Our evaluation protocol differs from the one used by Guo et al. \cite{Guo2015}, where different methods are evaluated on randomly selected but different splits of the same category, which may cause inaccurate comparisons.

\textbf{Results.} The performance of all methods at test time is reported in Table \ref{tab:results_per_category} for the ShapeNetCore dataset. The labeling accuracy for a given shape is measured as the percentage of surface points labeled correctly according to the ground-truth point labeling provided by Yi et al. \cite{Yi2016}. When considering a simple average of the per-category accuracies, our method performs $\mathbf{5.4\%}$ better than the best-performing prior work \cite{Kalogerakis2010} (Table \ref{tab:results_aggregated}, category average). Note, however, that several categories have disproportionately few models. A possibly more objective aggregate measure would be to weight each category by the number of test shapes. In this case, our method improves upon the state-of-the-art by $\mathbf{6.3\%}$ (Table \ref{tab:results_aggregated}, dataset average). Most importantly, our method has significantly higher performance in categories with complex objects, such as motor vehicles, aircraft, and furniture, where the labeling task is also more challenging. For categories with more than $3$ part labels, where part labeling is not just binary or ternary, our method improves upon prior work by $\mathbf{7.8\%}$ in the unweighted estimate (Table \ref{tab:results_aggregated}, category average, \textgreater3 labels), or by $\mathbf{7.9\%}$ when weighting by category size (Table \ref{tab:results_aggregated}, dataset average, \textgreater3 labels). This clearly indicates that our method can handle difficult shape labeling tasks in classes with complex objects significantly better than prior approaches. We include labeling results for all test shapes in the supplementary material.
Figure \ref{fig:gallery} demonstrates human-labeled (ground-truth) segmentations, along with results from the best performing prior method (ShapeBoost) and our method (ShapePFCN) for various test shapes. \rev{We found that ShapeBoost, which  relies on geometric descriptors, often fails for shapes with complex structure and topology (e.g. bikes, chairs), shapes with fine local features that  can distort local descriptors (e.g. gun trigger, baggage handles), and shapes with coarse geometry (e.g., hats).}

We also evaluated labeling accuracy in the PSB and COSEG datasets. We did not exclude any shape categories from our evaluation. Often, geometric methods are only applicable to certain types of ``well-formed'' input (e.g. manifolds, upright-oriented shapes, isometric articulations of a template, etc), and hence are not tested on unsuitable categories \cite{kai2015star}. Our method, by contrast, is broadly applicable. We obtain an improvement over state-of-the-art methods for these datasets ($92.6\%$ for our method, $90.6\%$ for ShapeBoost \cite{Kalogerakis2010}, $86.3\%$ for Guo et al. \cite{Guo2015} averaged over both datasets, see supplementary material for accuracy per category). \rev{We note that both PSB\ and COSEG\ contain a small number of shapes with limited variability, which even a shallow classifier may handle with high accuracy.}

\begin{figure}[t!]
  \begin{center}
  \includegraphics[width=0.84\columnwidth]{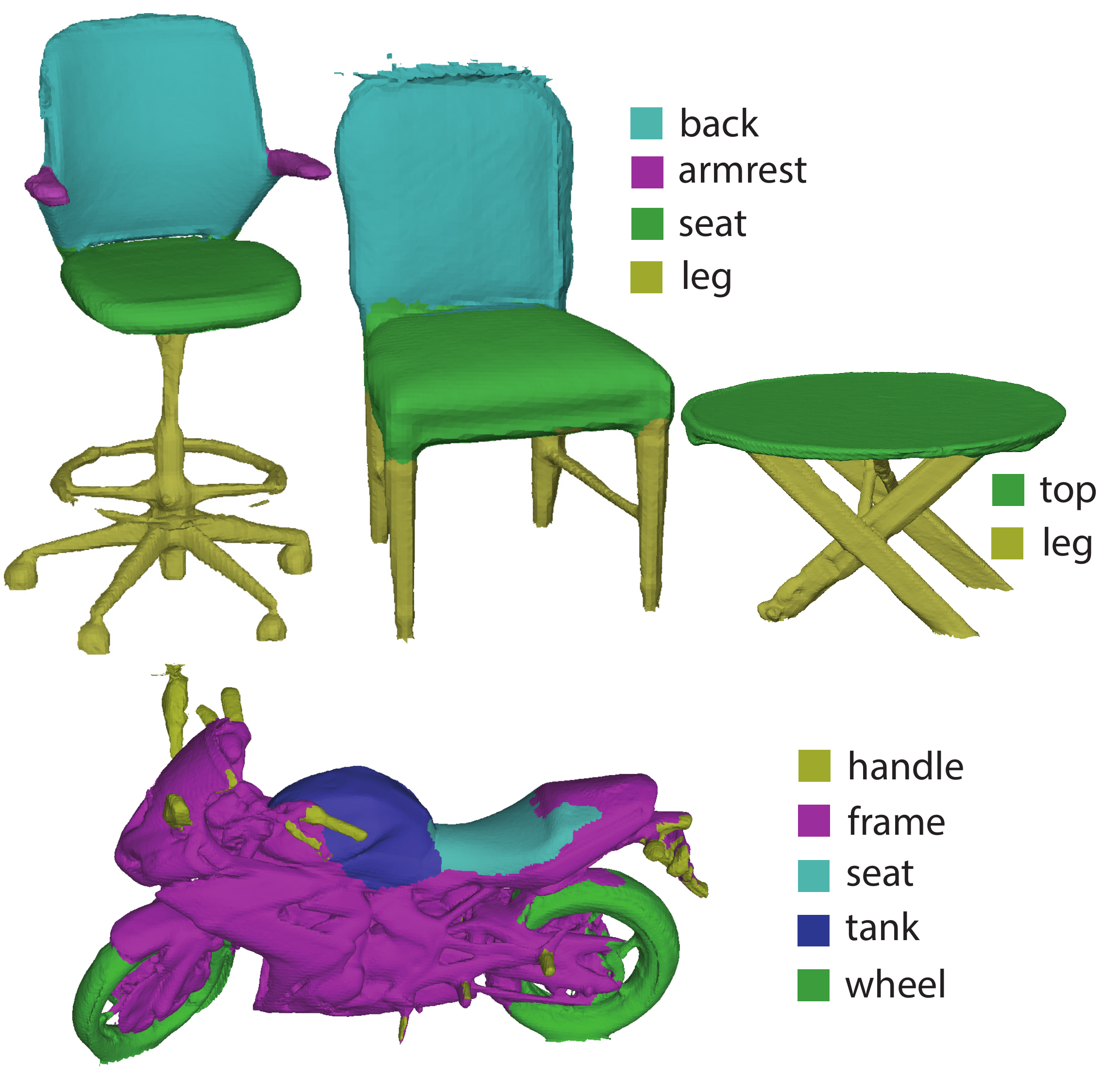}
\vskip -1mm
  \caption{\label{fig:scans} Labeled segmentations produced by our method on noisy objects reconstructed from RGBD sensor data.}
  \end{center}
\vskip -2mm
\end{figure}
\vspace{-1mm}

\begin{table}[t]
\small
  \centering
    \begin{tabular}{@{}c@{}|@{}c@{}|@{}c@{}|@{}c@{}|@{}c@{}|@{}c@{}}
    & \,fixed\, & \,disjoint\,    & \,unary\, & \,without\, & \,full\, \\
    & \,views\, & \,training\,    & \,term\, & \,pretrain.\, & \,method\, \\
    \hline
    \hline
    Category Avg. & 87.2 & 87.0 & 83.5 & 86.3 & \textbf{88.4} \\
    \hline
    Category Avg. (\textgreater3 labels)\, & 83.2 & 82.8 & 78.8& 82.5 & \textbf{85.0} \\
    \hline
    Dataset  Avg. & 86.2 & 85.9 & 82.1 & 85.7 & \textbf{87.5} \\
    \hline
    Dataset  Avg. (\textgreater3 labels) & 82.9 & 82.4 & 78.7 & 82.3 & \textbf{84.7 }\\
    \hline
    \end{tabular}%
  \caption{Labeling accuracy on ShapeNetCore for degraded variants of our method.}
\vspace{-4mm}
  \label{tab:results_alternatives}%
\end{table}%

\textbf{Analysis.} We also evaluated our method against alternative degraded variations of it, to identify major sources of performance gains. Table \ref{tab:results_alternatives} reports test labeling accuracy on ShapeNetCore for the following cases: (i) instead of selecting viewpoints that maximize surface coverage at different scales, we select fixed viewpoints placed on the vertices of a dodecahedron as suggested in \cite{su2015multi} for shape classification (see ``fixed views'' column), (ii) we train the FCN module and CRF separately (``disjoint training'' column), (iii) we do not use the CRF, i.e. we rely only on the unary term (``unary term'' column), (iv) we train the FCN module from scratch instead of starting its training from the pre-trained VGG.
We note that the CRF in particular, whether trained jointly or separately, is responsible for a major performance improvement (compare ``unary term'' to other columns). Pre-training also offers a noticeable gain. Viewpoint adaptation and joint training  contribute smaller but still useful gains.


\textbf{Generalization to RGB-D sensor data.} Even if our architecture is trained on complete, noise-free, manually-modeled 3D shapes, it can still generalize to noisy, potentially incomplete objects acquired from RGB-D sensors. Figure \ref{fig:scans} presents segmentation results for various objects originating from Choi et al.'s dataset \cite{choi2016large}. The dataset contains polygon meshes reconstructed from raw RGB-D sensor data. We trained our architecture on the ShapeNetCore chair, tables and motorbike categories (separately), then applied it to the reconstructed objects. We note that the scans included the ground that we removed heuristically through plane fitting. There was also background clutter that we removed through connected component analysis (i.e. we kept the dominant object in the scene).  In contrast to our method, prior works heavily rely on hand-coded geometric descriptors that are highly distorted by noisy, incomplete geometry and fail to produce meaningful results (see supplementary material for results on these objects).

\begin{table}[t]
\footnotesize
 \centering
    \begin{tabular}{@{}c@{}||@{}c@{}|@{}c@{}|@{}c@{}|@{}c@{}|@{}c@{}}
      & \,\#train/test\, & \#part & \,ShapePFCN\,    & \,ShapePFCN\,\\
      &     shapes       & \,labels\,& no upright    & upright     \\
    \hline
    \hline
    Airplane    & 250 / 250   & 4    & 90.3 & \textbf{91.2}  \\
    \hline
    Bag         & 38 / 38     & 2    & 94.6 & \textbf{94.9} \\
    \hline
    Cap         & 27 / 28     & 2    & \textbf{94.5} & 93.6 \\
    \hline
    Car         & 250 / 250   & 4    & 86.7 & \textbf{87.5}  \\
    \hline
    Chair       & 250 / 250   & 4    & 82.9 & \textbf{85.5} \\
    \hline
    Earphone    & 34 / 35     & 3    & 84.9 & \textbf{85.6} \\
    \hline
    Guitar      & 250 / 250   & 3    & 91.8 & \textbf{92.5} \\
    \hline
    Knife       & 196 / 196   & 2    & 82.8 & \textbf{83.8} \\
    \hline
    Lamp        & 250 / 250   & 4    & 78.0 & \textbf{81.3}  \\
    \hline
    Laptop      & 222 / 223   & 2    & \textbf{95.3} & 95.1 \\
    \hline
    Motorbike   & 101 / 101   & 6    & 87.0 & \textbf{87.4} \\
    \hline
    Mug         & 92 / 92     & 2    & \textbf{96.0} & 95.9 \\
    \hline
    Pistol      & 137 / 138   & 3    & \textbf{91.5} & 91.4  \\
    \hline
    Rocket      & 33 / 33     & 3    & 81.6 & \textbf{83.9} \\
    \hline
    Skateboard  & 76 / 76     & 3    & 91.9 & \textbf{92.1} \\
    \hline
    Table      & 250 / 250    & 3    & 84.8 & \textbf{87.8} \\
    \hline
    \hline
    Category Avg.                          & \multicolumn{2}{c}{} & 88.4 & \textbf{89.4} \\
    \hline
    Category Avg. (\textgreater3 labels)\, & \multicolumn{2}{c}{} & 85.0 & \textbf{86.6} \\
    \hline
    Dataset  Avg.                          & \multicolumn{2}{c}{} & 87.5 & \textbf{88.8} \\
    \hline
    Dataset  Avg. (\textgreater3 labels)   & \multicolumn{2}{c}{} & 84.7 & \textbf{86.5 }\\
    \hline
    \end{tabular}%
  \caption{Dataset statistics, labeling accuracy per category, and aggregate labeling accuracy for our test split in the case of consistent upright shape orientation and additional input channel in our rendered images for encoding the  upright axis coordinate values (height from the ground plane). The\ labeling accuracy of our network is improved for most classes and on average.}
\vspace{-4mm}
  \label{tab:upresults_per_category}
\end{table}%

\vspace{-2mm}
\section{Conclusion}
\vspace{-3mm}
We presented a deep architecture designed to segment and label 3D shape parts. The key idea of our approach is to combine image-based fully convolutional networks for view-based reasoning, with a surface-based projection layer that aggregates FCN outputs across multiple views and a surface-based CRF to favor coherent shape segmentations. Our method significantly outperforms prior work on 3D shape segmentation and labeling.

There are several exciting avenues for future extensions. Currently our method uses a simple pairwise term based on surface distances and angles between surface normals. As a result, the segmentations can become noisy and not aligned with strong underlying mesh boundaries (Figure \ref{fig:scans}, see motorbike). Extracting robust  boundaries through a learned module would be beneficial to our method. Our method currently deals with single-level, non-hierarchical segmentations. Further segmenting objects into fine-grained parts (e.g. segmenting motorbikes into sub-frame components) in a hierarchical manner would be useful in several vision and graphics applications. \rev{Another possibility for future work is to investigate different types of input to our network. 
\footnote{After the initial publication of our work, we also incorporated a third channel in our  rendered images representing height from ground plane. This setting should be used only in the case of given consistent upright orientation for all input training and test shapes. Although consistent upright orientation is not true for 3D\ models in general, ShapeNetCore provides it and several researchers have already tested their methods under the assumption of consistent upright or even fully consistent orientation. We include  labeling accuracy based on our \ training and test splits in Table \ref{tab:upresults_per_category}  in the case of consistent upright orientation and  with this new input channel.  } 
The  input images we used represent surface depth and normals relative to  view direction. Another possibility is to consider the HHA encoding \cite{gupta2013perceptual} or even raw position data. However, these encodings assume a consistent gravity direction or alignment for input 3D shapes. Although in a few repositories (e.g. Trimble Warehouse) the majority of 3D models have consistent upright orientation, this does not hold for all 3D\ models, and especially for other online repositories whose shapes are oriented along different, random axes. There have been efforts to develop methods for consistent orientation or alignment of 3D\ shapes \cite{Fu:2008,gupta2013perceptual,Chang2015}, yet existing methods require human supervision, or do not work well for various shape classes, such as outdoor objects or organic shapes.}

Finally, our method is currently trained in a fully supervised manner. Extending our architecture to the semi-supervised or unsupervised setting to benefit from larger amounts of data is another exciting future direction.

\paragraph{Acknowledgements.}
\rev{Kalogerakis acknowledges support from NSF (CHS-1422441, CHS-1617333), NVidia and Adobe. Maji acknowledges support from NSF (IIS-1617917) and Facebook. Chaudhuri acknowledges support from Adobe and Qualcomm. Our experiments were performed in the  UMass GPU cluster  obtained under a grant from the Collaborative R\&D Fund managed by the Massachusetts Technology Collaborative.}

\appendix
\section{Supplementary Material}

\subsection{Evaluation in PSB/COSEG}

The labeling accuracy of our method (ShapePFCN), ShapeBoost \cite{Kalogerakis2010} and Guo et al. \cite{Guo2015} per category is presented in Table \ref{tab:psbcoseg_per_category}. Aggragate performance is shown in Table \ref{tab:psbcoseg_aggregated}.
 The labeling accuracy for a  shape is measured as the percentage
of surface area labeled correctly according to the ground-truth face labeling provided in the L-PSB  \cite{Kalogerakis2010} and COSEG \cite{Wang2012} datasets. 

\subsection{ShapeBoost results on RGB-D sensor data}

We applied ShapeBoost on the same objects used in Figure 4 of our paper. The method failed to produce compelling results - see Figure \ref{fig:scans2} below, and compare with the results of our method shown in Figure 4 of our paper. We suspect that the underlying reason for these failure cases of ShapeBoost (and in general methods that rely on hand-engineered geometric descriptors) is that noise, holes, and mesh degeneracies easily distort geometric descriptors. Another potential reason is that shallow classifiers tend to underfit datasets of shapes with significant variability.

\begin{figure}[h]
  \begin{center}
  \includegraphics[width=0.8\columnwidth]{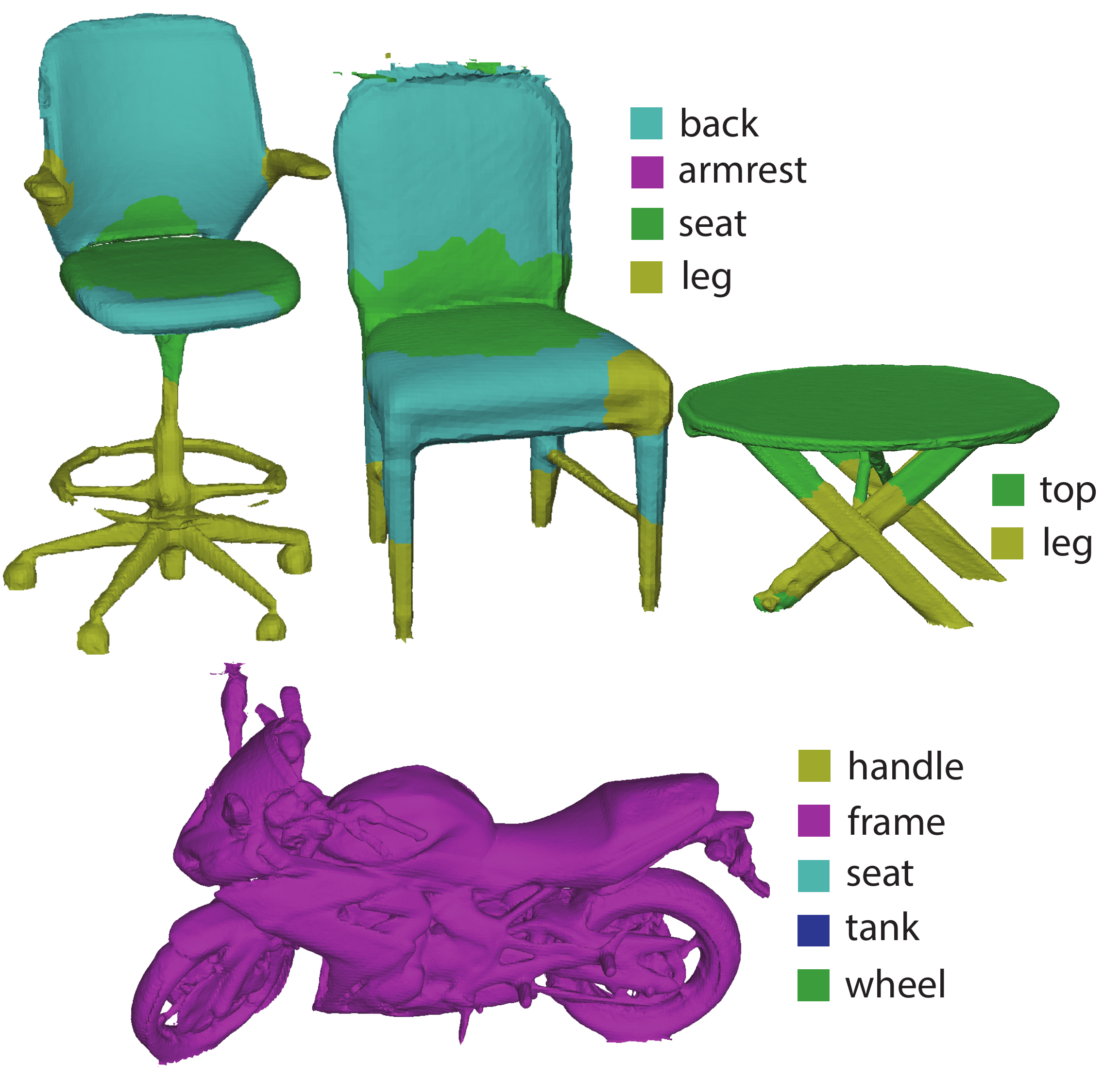}
  \caption{\label{fig:scans2} Labeled segmentations produced by ShapeBoost on noisy objects reconstructed from RGBD sensor data.}
  \end{center}
\end{figure}

\subsection{Additional data}

We provide visualizations of segmentations produced by our method, ShapeBoost \cite{Kalogerakis2010} and Guo et al. \cite{Guo2015} on our test shapes 
from ShapeNetCore, PSB\ and COSEG
in our project page (see: http://people.cs.umass.edu/kalo/papers/shapepfcn/). We also provide a text file (\textit{splits.txt}) that includes the training and test splits we used in our experiments.

\begin{table}[h]
\footnotesize
 \centering
    \begin{tabular}{@{}c@{}||@{}c@{}|@{}c@{}|@{}c@{}|@{}c@{}|@{}c@{}}
      & \,\#train/test\, & \#part& \multirow{2}{*}{\,ShapeBoost\,} & \multirow{2}{*}{\,Guo et al.\,}    & \multirow{2}{*}{\,ShapePFCN\,}\\
      &     shapes       & \,labels\,&                          &                            &   \\
    \hline
    \hline
    psbAirplane & 12 / 8 & 5     & 96.1  & 91.6  & 93.0 \\
    \hline
    psbAnt & 12 / 8 & 5     & 98.7  & 97.6  & 98.6 \\
    \hline    
    psbArmadillo & 12 / 8 & 11    & 92.6  & 85.0  & 92.8 \\
    \hline        
    psbBearing & 12 / 8 & 5     & 92.2  & 77.4  & 92.3 \\
    \hline        
    psbBird & 12 / 8 & 5     & 89.6  & 83.1  & 88.5\\
    \hline        
    psbBust & 12 / 8 & 8     & 63.4  & 34.8  & 68.4 \\
    \hline        
    psbChair & 12 / 8 & 4     & 98.1  & 96.7  & 98.5 \\
    \hline        
    psbCup & 12 / 8 & 2     & 94.0  & 92.1  & 93.8 \\
    \hline        
    psbFish & 12 / 8 & 3     & 95.7  & 94.5  & 96.0 \\
    \hline        
    psbFourLeg & 12 / 8 & 6     & 83.3  & 82.4  & 85.0 \\
    \hline        
    psbGlasses & 12 / 8 & 3     & 96.9  & 95.3  & 96.6 \\
    \hline        
    psbHand & 12 / 8 & 6     & 94.4  & 73.8  & 84.8 \\
    \hline        
    psbHuman & 12 / 8 & 8     & 86.8  & 85.6  & 94.5 \\
    \hline        
    psbMech & 12 / 8 & 5     & 99.5  & 98.5  & 98.7 \\
    \hline        
    psbOctopus & 12 / 8 & 2     & 98.2  & 97.4  & 98.3 \\
    \hline        
    psbPlier & 12 / 8 & 3     & 95.2  & 95.2  & 95.5 \\
    \hline        
    psbTable & 12 / 8 & 2     & 99.4  & 98.5  & 99.5 \\
    \hline        
    psbTeddy & 12 / 8 & 5     & 98.7  & 97.3  & 97.7 \\
    \hline        
    psbVase & 12 / 8 & 5     & 81.7  & 77.8  & 86.8 \\
    \hline        
    cosegCandelabra & 12 / 16 & 4     & 85.5  & 85.9  & 95.4 \\
    \hline        
    cosegChairs & 12 / 8 & 3     & 94.8  & 93.8  & 96.1 \\
    \hline        
    cosegFourleg & 12 / 8 & 5     & 92.3  & 88.2  & 90.4 \\
    \hline        
    cosegGoblets & 6 / 6 & 3     & 97.0  & 86.1  & 97.2 \\
    \hline        
    cosegGuitars & 12 / 32 & 3     & 97.7  & 97.7  & 98.0 \\
    \hline        
    cosegIrons & 12 / 6 & 3     & 87.2  & 79.7  & 88.0 \\
    \hline        
    cosegLamps & 12 / 8 & 3     & 76.3  & 78.0  & 93.0 \\
    \hline        
    cosegVases & 12 / 16 & 4     & 86.4  & 84.4  & 84.8 \\
    \hline        
    cosegVasesLarge & 12 / 288 & 4     & 89.7  & 80.1  & 90.6 \\
    \hline        
    cosegChairsLarge & 12 / 388 & 3     & 76.5  & 80.8  & 91.1 \\
    \hline        
    cosegTeleAliens & 12 / 188 & 4     & 81.7  & 80.0  & 95.7 \\    
    \hline
    \end{tabular}%
  \caption{Dataset statistics and labeling accuracy per category for test shapes in PSB \&\ COSEG.}
  \label{tab:psbcoseg_per_category}
\end{table}%

\begin{table}[h]
\small
  \centering
    \begin{tabular}{@{}c@{}|@{}c@{}|@{}c@{}|@{}c@{}}
    & \,ShapeBoost\, & \,Guo et al.\,    & \,ShapePFCN\, \\
    \hline
    \hline
    Category Avg. & 90.6 & 86.3 & \textbf{92.6} \\
    \hline
    Category Avg. (\textgreater3 labels)\, & 89.5 & 83.3 & \textbf{90.9} \\
    \hline
    Dataset  Avg. & 84.2 & 82.1 & \textbf{92.2} \\
    \hline
    Dataset  Avg. (\textgreater3 labels) & 87.2 & 81.0 & \textbf{92.1 }\\
    \hline
    \end{tabular}%
  \caption{Aggregate labeling accuracy on  PSB \&\ COSEG.}
  \label{tab:psbcoseg_aggregated}
\end{table}%

\clearpage

{\small
\bibliographystyle{ieee}
\bibliography{egbib}
}

\end{document}